\documentclass{article}
\usepackage[utf8]{inputenc}
\usepackage[margin=1in]{geometry}
\usepackage[dvipsnames]{xcolor}
\usepackage{xspace}
\usepackage{url}

\usepackage[utf8]{inputenc}
\usepackage[T1]{fontenc}

\usepackage{graphicx}
\usepackage{dcolumn}% Align table columns on decimal point
\usepackage{bm}% bold math
\usepackage{amsmath}
\usepackage{comment} 
\usepackage{amssymb}
\usepackage{mathrsfs} %jolie lettres
\usepackage[normalem]{ulem}

\usepackage{float} 
\usepackage{hyperref}
\usepackage{algorithm}
\usepackage{algorithmic}
\usepackage[caption=false]{subfig}

\usepackage{authblk}

\title{Message Passing Descent for Efficient Machine Learning}
\author{Francesco Concetti \\ e-mail: \href{mailto:concetti@math.arizona.edu}{concetti@math.arizona.edu}
   \and Misha Chertkov \\ e-mail:  \href{mailto:chertkov@arizona.edu}{chertkov@arizona.edu} }
\affil{Program in Applied Mathematics, University of Arizona, Tucson, USA}
\date{\today}

\begin{document}

\maketitle

%\maketitle
\begin{abstract} We propose a new iterative optimization method for the {\bf Data-Fitting} (DF) problem in Machine Learning, e.g. Neural Network (NN) training. The approach relies on {\bf Graphical Model} (GM) representation of the DF problem, where variables are fitting parameters and factors are associated with the Input-Output (IO) data. The GM results in the {\bf Belief Propagation} Equations considered in the {\bf Large Deviation Limit} corresponding to the practically important case when the number of the IO samples is much larger than the number of the fitting parameters. We suggest the {\bf Message Passage Descent} algorithm which relies on the piece-wise-polynomial representation of the model DF function. In contrast with the popular gradient descent and related algorithms our MPD algorithm rely on analytic (not automatic) differentiation, while also (and most importantly) it descents through the rugged DF landscape by \emph{making non local updates of the parameters} at each iteration. The non-locality guarantees that the MPD is not trapped in the local-minima, therefore resulting in better performance than locally-updated algorithms of the gradient-descent type. We illustrate  superior performance of the algorithm on a Feed-Forward NN with a single hidden layer and  a piece-wise-linear activation function.   
\end{abstract}

\vspace{-0.7cm}
\section{Introduction}\label{sec:intro}  

\vspace{-0.2cm}
Fitting models to data is in the core of the Machine Learning (ML). The models may take a form of a Neural Networks (NN) or the model may be informed by an application, e.g. physics.  Either way the fitting models are expected to have enough of parameters, usually real valued,  which allow sufficient flexibility in fitting the data. Given that the class of the parameterized Model Functions (MF) is fixed,  the next step in posing the data fitting problem becomes to select the Loss Function (LF) minimizing over the parameters the mismatch between the data and the prediction of the MF.  This manuscript focuses on devising an efficient algorithm for solving the data fitting optimization problem in the setting where the data and the model function are fixed. Strategically, we are interested in addressing the most challenging cases, where the \emph{resulting multivariate LF landscape is rugged}, such that most popular ML algorithms, searching for the minimum in the LF landscape through a sequence of local steps, either fail or underperform. 

\vspace{-0.1cm}
We suggest a novel approach which marches through the rugged landscape via a sequence of large,  that is not incremental, steps. The approach consists in restating, in Section \ref{sec:formulation}, the LF optimization problem in a  high dimensional, functional, space , representing proxies for the probability distribution functions of the parameters, also called beliefs. This functional optimization can be interpreted as minimization of the Gibbs-Kullback-Leibler (GKL) measure in the functional space of beliefs (belief functions). The GKL measure compares beliefs with the probability distribution factorized through a Graphical Model (GM) in a bi-partite Factor-Graph representation, where factors are associated with samples (respective contribution into the LF) and nodes are associated with variables. We apply the Bethe approach \cite{2005Yedidia} to approximate the functional GKL optimization,  therefore arriving at the Belief Propagation (BP) Equations for the message functions (which are functions conjugated to the belief functions).  The BP equations are functional and integral,  i.e. non-local in the space of parameters.  We conjecture (but do not prove) that the Bethe approximation, and therefore BP equations, become exact in the limit where number of samples and number of parameters is asymptotically large. Furthermore,  if the number of samples is also significantly larger than the number of parameters (data fitting regime of practical inference avoiding undesirable over-fitting) the BP equations simplify to the reduced BP equations which we solve iteratively with the Message Passing Descent (MPD) algorithm, introduced in Section \ref{sec:MPD-derivation}. Most importantly, and in spite of the simplification, MPD still requires solving a sequence of global optimization problems, each optimizing a message function dependent on a single parameter,  where other parameters remain fixed to their current values.  This non-locality in the space of parameters make the MPD algorithm principally different from the parameter-space local algorithms,  e.g. any of the algorithm from the Gradient Descent (GD) family popular in Deep Learning (DL). 

\vspace{-0.1cm}
In other striking deviation from the mainstream approaches popular in ML of today, we show how to resolve each step of the MPD analytically which allows to avoid using automatic differentiation and leads to efficient implementation in the case when the MFs of the fitting problem are Piece-Wise-Polynomial (PWP).  We illustrate, in Section \ref{sec:experiments}, utility of our approach on the examples of the Feed-Forward Neural Networks (FF-NNs) with one hidden layer, leaky version of the hard-tanh activation function and the $L_2$-norm LF applied to data from \cite{UCI} database with sufficiently rugged LF landscapes.  We observe that, as predicted, the MPD outperforms Adam and Nesterov-Accelerated Gradient (two most popular training algorithm of the GD type).

%\misha{... start by explaining the niche this manuscript fits in --- design of new algorithms improving upon existing ... have a review of existing methods/algorithms,  e.g. Gradient Descent (GD) and related (main ideas and references - briefly) ... emphasize generality of the approach ... new take, which does not rely on automatic differentiation ... utilizing analytical structure as much as possible ... }

% Gradient Descent is the most common optimization algorithm in machine learning and deep learning. It is a first-order optimization algorithm. This means it only takes into account the first derivative when performing the updates on the parameters. On each iteration, we update the parameters in the opposite direction of the gradient of the objective function J(w) w.r.t the parameters where the gradient gives the direction of the steepest ascent. The size of the step we take on each iteration to reach the local minimum is determined by the learning rate α. Therefore, we follow the direction of the slope downhill until we reach a local minimum.

\vspace{-0.3cm}
\section{Problem Formulation: Data Fitting}\label{sec:formulation}
\vspace{-0.2cm}

In this Section, we first present a Graphical Model (GM) representation for the parametric Data Fitting (DF) problem and then describe a Message-Passing (MP) algorithm solving the problem.

\vspace{-0.2cm}
\subsection{Parametric Data Fitting}\label{sec:data-fitting}
\vspace{-0.2cm}

Consider a system which takes a $d_{\text{in}}$-dimensional input $\bm{x}=(x_d|d\in [d_{\text{in}}])\in \mathcal{X}\subset \mathbb{R}^{d_{\text{in}}}$ and maps it into an $d_{\text{out}}$-dimensional output, $\bm{y}=(y_d|d\in [d_{\text{out}}])\in \mathcal{Y}\subset \mathbb{R}^{d_{\text{out}}}$. We use the short-hand notation $[d_{\text{in}}]\doteq \{1,\cdots,d_{\text{in}}\}$. Our observation of the system is represented via sufficiently large, $S\gg 1$, number of the Input-Output (IO) samples
\begin{gather}\label{eq:IO-N}
\mathbb{IO}^{(S)}=\{(\bm{x}^{(s)},\bm{y}^{(s)})\left| s\in [S]\right.\} \subset \left(\mathcal{X}\times \mathcal {Y}\right)^S\,.
\end{gather}
The aim of the supervised Machine Learning is to reconstruct the map $\bm{x}\mapsto \bm{y}$ from the data-set $\mathbb{IO}^{(S)}$ defined in Eq.~(\ref{eq:IO-N}). In general,  the reconstruction  is posed by considering a class of the parameterized Model Functions (MF)
\begin{equation}
\label{eq:g-model}
g[\bm{\theta}]:\mathcal{X}\longrightarrow \mathcal{Y}\,,
\end{equation}
where $\bm{\theta}$ is the $P$-dimensional vector of \textbf{parameters}, and a learning objective, called the \textbf{Loss Function} (LF):
\begin{equation}
\label{eq:loss_fun}
L_{\mathbb{IO}^{(S)}}(\bm{\theta})\doteq \frac{1}{S}\sum^S_{s=1} l(\bm{x}^{(s)},\bm{y}^{(s)},\bm{\theta}),\ \bm{\theta}\in \mathbb{R}^{P}\,.
\end{equation}
The \textbf{log-likelihood} function $l$ compares the output prediction derived from the input sample ${\bm x}$, according to the MF $g_{\bm \theta}$, with the output sample, ${\bm y}$
\begin{gather}\label{eq:LL}
%l:\mathcal{X}\times \mathcal{Y}\times \mathbb{R}^P\longrightarrow \mathbb{R}_+,\quad 
l(\bm{x},\bm{y} ,\bm{\theta})=\omega(\bm{g}_{\bm\theta}(\bm{x}),\bm{y}),
\end{gather}
where $\omega:\mathcal{Y}^2\longrightarrow \mathbb{R}_+$ is a positive scalar function which is usually chosen to be convex with respect to its first argument. In particular, we provide a numerical test for the mean square error loss-function:
\begin{equation}
\label{eq:LL-l2}
l_2(\bm{x},\bm{y} ,\bm{\theta}) =\|\bm{y}-\bm{g}_{\bm\theta}(\bm{x})\|^2=\sum^{d_{\text{out}}}_{d=1}|y_d-g_{d;\bm{\theta}}({\bm x})|^2.
\end{equation}
Given the $\mathbb{IO}^{(S)}$ data set, optimal reconstruction is achieved by minimizing the LF (\ref{eq:loss_fun}) over the vector of parameters
\begin{equation}
\label{eq:training}
\bm{\theta}^*_{\mathbb{IO}^{(S)}}\doteq\underset{\bm{\theta}\in\mathbb{R}^P}{\text{argmin}}\,\,L_{\mathbb{IO}^{(S)}}(\bm{\theta}).
\end{equation}
%The loss-function, usually, is not convex and it may present several local minima.
The main idea of our approach consists in \emph{restating the general data-fitting problem as a Maximum Likelihood of a Graphical Model (GM)}.

\vspace{-0.2cm}
\subsection{Data Fitting as a Graphical Model}\label{sec:GM}
\vspace{-0.2cm}

\begin{figure}
    \centering
  \includegraphics[width=0.5\textwidth]{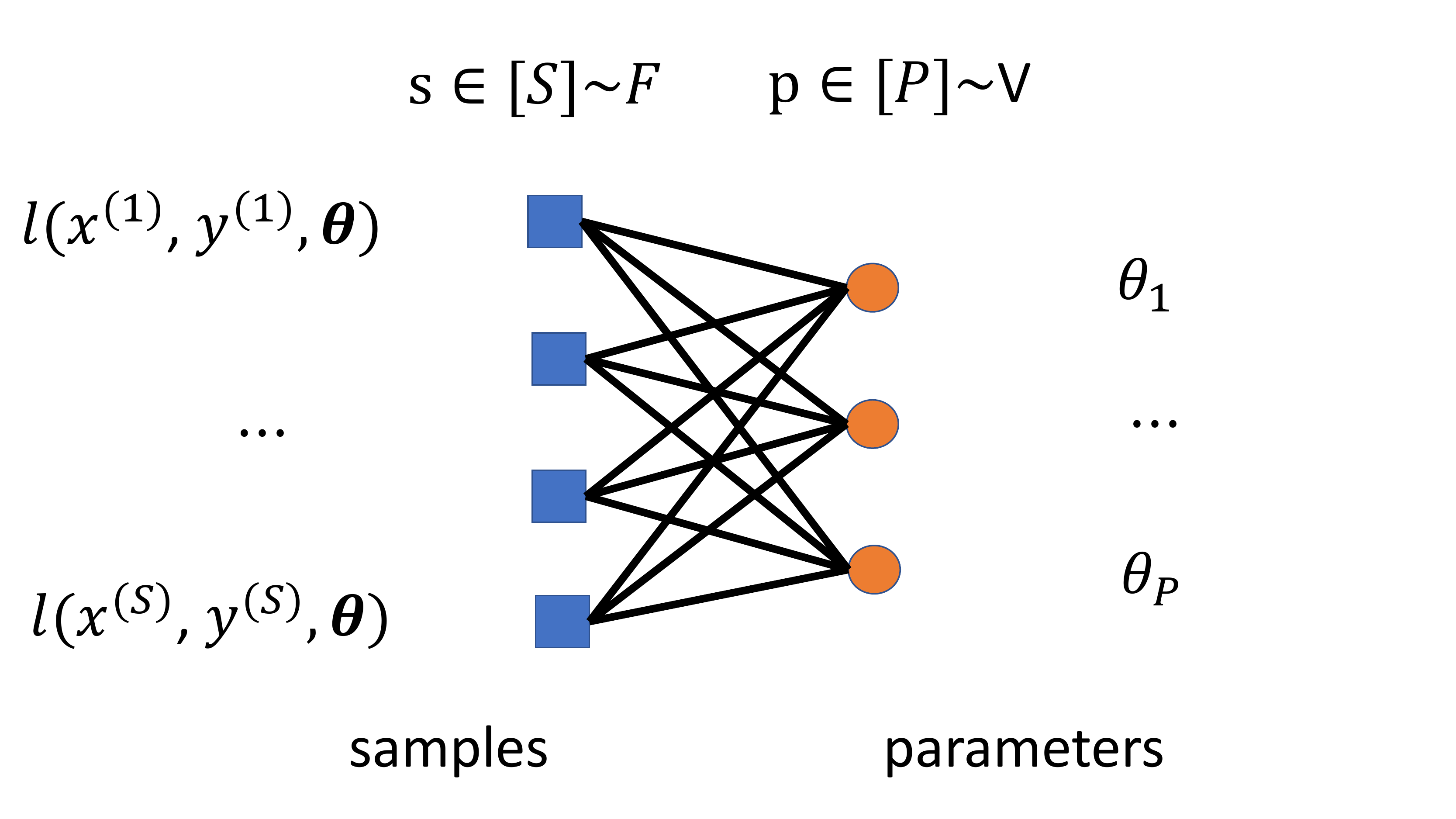}\vspace{-0.5cm} 
    \caption{Graphical Model representing the Data Fitting problem. Factors and Nodes are associated with samples and parameters respectively.}
    \label{fig:FG}
\end{figure}

Let $\mathcal{G}$ be a fully connected bi-partite factor graph, with $P$ vertices, $S$ factors and $S\times P$ edges connecting vertices and factors. (See Fig.~(\ref{fig:FG}).) We denote the set of vertices by $\mathcal{V}$, the set of factors by $\mathcal{F}$ and the set of edges by $\mathcal{E}$. Each IO sample, $(\bm{x}^{(s)},\bm{y}^{(s)})\in \mathbb{IO}^{(S)}$, is associated with a factor $s\in [S]\sim \mathcal{F}$, and each parameter $\theta_p$ which is a component of the vector of parameters $\bm{\theta}$ is associated with a vertex $p\in [P]\sim \mathcal{V}$. The graph $\mathcal{G}$ encodes a conditional probability density of the parameters $\bm{\theta}$, given the IO sample:
\begin{gather}
\label{eq:GM-prob}
p_{\beta}(\bm{\theta}|\mathbb{IO}^{(S)}) =\frac{1}{Z(\beta,\mathbb{IO}^{(S)})} e^{-\beta S\,L_{\mathbb{IO}^{(S)}}(\bm{\theta})}
%\\ \nonumber & =\frac{1}{Z(\beta,\mathbb{IO}^{(N)})} \prod_{s\in \mathcal{F}}e^{-\beta l(\bm{x}^{(s)},\bm{y}^{(s)},\bm{\theta})}\,,
\end{gather}
where $\beta$ is a positive regulation parameter and $Z(\beta,\mathbb{IO}^{(S)})$ is the normalization constant, called the Partition Function (PF).
The $\beta$-regulation is motivated by statistical physics, where $\beta$ has the meaning of the inverse temperature and the normalization constant,  $Z(\beta,\mathbb{IO}^{(S)})$, is also called the Partition Function (PF). The LF optimization (\ref{eq:training}) allows the following Maximum Likelihood (ML) reformulation
\begin{equation}
\label{eq:ML}  \bm{\theta}^*_{\mathbb{IO}^{(S)}} = \underset{\bm{\theta}\in\mathbb{R}^P}{\text{argmax}}\ p_{\beta}(\bm{\theta}|\mathbb{IO}^{(S)})
\end{equation}
Continuing with the statistical physics analogy (and intuition), in what follows the loss-function, $L_{\mathbb{IO}^{(S)}}(\bm{\theta})$,  plays the roll of the model Hamiltonian; the distribution, $p_{\beta}(\bm{\theta}|\mathbb{IO}^{(S)})$, is the Gibbs distribution.
Solution of \eqref{eq:training}, $\bm{\theta}_{\mathbb{IO}^{(S)}}^{*}$, is the ground state configuration of the model, and $L_{\mathbb{IO}^{(S)}}(\bm{\theta}_{\mathbb{IO}^{(S)}}^{*})$ is the ground state energy. At zero temperature, $\beta\to\infty$, the distribution $p_{\beta}(\bm{\theta}|\mathbb{IO}^{(S)})$ concentrates around the ground state configuration. 
\vspace{-0.2cm}
\subsection{Bethe Free Energy, Belief Propagation Equations and Message Passing Algorithm}\label{sec:BP}\vspace{-0.2cm}

Computing log-PF of the GM (\ref{eq:GM-prob}) can be re-stated as the (Gibbs/Kullback-Leibler) functional optimization
\begin{gather} \label{eq:Gibbs}
    %\log Z(\beta,\mathbb{IO}^{(S)})=
    \min\limits_{\{{\cal B}\}} \int d{\bm\theta} {\cal B}({\bm\theta})\log\left(\frac{\exp\left(-\beta S\,L_{\mathbb{IO}^{(S)}}(\bm{\theta})\right)}{{\cal B}({\bm\theta})}\right),
\end{gather}
over beliefs, $\{{\cal B}\}=({\cal B}({\bm \theta})\geq 0|\forall {\bm\theta}; \int d{\bm\theta} {\cal B}({\bm\theta})=1)$, which are proxies for probabilities \footnote{Notice in passing that the GKL reformulation (\ref{eq:Gibbs}) of the DF problem is also a starting point for the general purpose Probability Functional Descent (PFD) algorithm introduced in \cite{2019Jose}.  It was shown that PFD is a high-level blueprint for many other algorithms recently discussed in the context of DL. In this regards, the MPD algorithm -- major invention of this manuscript introduced in the following - may also be considered as a very special implementation of the PFD algorithm, taking advantage of the GM structure of the DF formulation.}. Under the ansatz that the belief, ${\cal B}({\bm \theta})$, can be expressed in terms of its single parameter marginals, $\forall p\in [P]:\ {\cal B}_p(\theta_p)\doteq \int_{{\bm\theta}\setminus \theta_p} {\cal B}({\bm\theta})$, according to 
\begin{gather*}
    {\cal B}({\bm \theta})\to ({\cal B}({\bm \theta}))^{S} \prod\limits_{p\in [P]}\left({\cal B}_p(\theta_p)\right)^{1-S},
\end{gather*}
the optimization (\ref{eq:Gibbs}) is transformed into its approximate version,  called the Bethe Free Energy (BFE) functional \cite{2005Yedidia}. The optimization set by the BFE functional (not shown here due to space limitations) is exact if the underlying graph is a tree. We conjecture,  following the general logic wide-spread in the statistical physics literature,  see e.g. \cite{VPM} and references there in, that in the limit where the number of samples and the number of parameters become infinite, the BFE functional approach become asymptotically exact \footnote{Focused intentionally and primarily on the applied, algorithmic aspects of the novel formulation,  we do not attempt to prove the conjecture in our first manuscript on the subject. However, we consider the largely theoretical subject important and plan to address it in future publications.}. In general, BFE approach is expected to provide powerful and accurate heuristics when the graph is sufficiently large. 

Following the approach pioneered in \cite{2005Yedidia}, we resolve the BFE optimization via an iterative algorithm called Message-Passing (MP). For a given factor graph $\mathcal{G}$, the MP algorithm introduces a set of auxiliary variables, called messages, associated to the edges of the graph. Messages are Lagrangian multipliers (dual variables) for the beliefs' marginalization conditions. Euler-Lagrange equations for the messages, looking for the extrema of the BFE functionals, are called the Belief Propagation (BP) equations. Adapted to our (fully connected bi-partite) setting the BP equations for the messages, from vertices (parameters) to factors (samples), $\{m_{p\to s}(\theta_p);p\in \mathcal{V},\,s \in \mathcal{F}\}$, and from factors to vertices $\{m_{s\to p}(\theta_p);p\in \mathcal{V},\,s \in \mathcal{F}\}$, become 
\begin{align}
\label{eq:BP_factor_1_NN}
  & m_{s\to p}(\theta_p)\propto \\ \nonumber & \int d{\bm\theta} e^{-\beta  l(\bm{x}^{(s)},\bm{y}^{(s)},\bm{\theta})} \left(\prod_{p'\in\mathcal{V}/\{p\}} m_{p'\to s}(\theta_{p'})d\theta_{p'}\right),\\
\label{eq:BP_factor_2_NN}
& m_{p\to s}(\theta_p)  \propto \prod_{s'\in\mathcal{F}/\{s\}}m_{s'\to p}(\theta_p).
\end{align}
%We consider the messages $m_{p\to s}$ and $m_{p\to s}$ normalized, for all $(s,p) \in{\cal E}$.
Observe that the BP equations are factor-graph-local which suggests solution via passing messages (thus the name MP for the algorithm) from factors to nodes and back till convergence.  
%Eventually, the equations are solved sequentially, in such a way that the computation tree runs down through the graph. This is the main conceptual difference between MP methods and Newton-like methods, where all the unknown of the problems are updated in parallel toward a gradually better approximation of the solution.
%The BP is a MP algorithm that, when it converges, provides an approximation of the marginal distributions over the factor graph $\mathcal{G}$.
%The messages,  understood as approximations for the conditional probability distributions associated with directed edges set as follows: from vertices (parameters) to factors (samples), $\{m_{p\to s}(\theta_p);p\in \mathcal{V},\,s \in \mathcal{F}\}$, and from factors to vertices $\{m_{s\to p}(\theta_p);p\in \mathcal{V},\,s \in \mathcal{F}\}$.
%Formally, the BP equations for the GM \eqref{eq:GM-prob} are easy to derive, regardless of how complicated is the loss-function and the model function are. For any $s\in \mathcal{F}$ and $p\in \mathcal{V}$, the BP messages satisfy
The marginal probability of the parameter $\theta_p$, for any $p\in \mathcal{V}$, within the BFE approximation, is reconstructed from messages according to 
\begin{align}
\label{eq:BP_factor_dist}
 p_{p}(\theta_p|\mathbb{IO}^{(S)}) & \propto \prod_{s'\in\mathcal{F}}m_{s'\to p}(\theta_p),\\
\label{eq:BP_factor_dist_max}
 \theta^*_p & = \underset{\bm{\theta}\in\mathbb{R}^P}{\text{argmax}}\,\,p_{p}(\theta_p|\mathbb{IO}^{(S)}). %\\ \nonumber & = \lim_{\beta \to \infty}\int d\theta_p \, \theta_p p_{p\to s}(\theta_p|\mathbb{IO}^{(S)}).
\end{align}
Notice that Eqs.~(\ref{eq:BP_factor_1_NN},\ref{eq:BP_factor_2_NN}) are \textbf{functional} and also \textbf{integral} equations (not algebraic) and the main computational overhead, if we attempt to solve the equations directly,  is due to the integration over $|\mathcal{V}|-1=P-1$ variables in Eq.~(\ref{eq:BP_factor_1_NN}). If $P$ is large, the integration is computationally expensive therefore making the naive approach impractical. Even though we are mainly interested in nonlinear MF, $g_{\bm{\theta}}(\bm{x})$, it is appropriate to mention that in the special case of the linear in $\bm{\theta}$ model function  the MP Eqs.~(\ref{eq:BP_factor_1_NN},\ref{eq:BP_factor_2_NN}) become the so-called Generalized Approximated-Message-Passing (GAMP) equations discussed, e.g., in \cite{Rangar,Javanmard-Montanari}. 

\vspace{-0.4cm}
\section{Message Passing Descent Algorithm}\label{sec:MPD-derivation}
\vspace{-0.1cm}

In this Section we show how to simplify the BP Eqs.~(\ref{eq:BP_factor_1_NN},\ref{eq:BP_factor_2_NN}), therefore turning them into a practical algorithm which we coin the Message Passing Descent (MPD) algorithm. We achieve the goal in two steps. First, in Section \ref{sec:functional-algebraic} we reduce the BP Eqs.~(\ref{eq:BP_factor_1_NN},\ref{eq:BP_factor_2_NN}), which depend on  high-dimensional integrations, to
Eqs.~(\ref{eq:u_BP},\ref{eq:h_BP},\ref{eq:global_min_BP}), coined reduced-BP equations, where the high-dimensional integrations are replaced by optimizations of functions of a single parameter. Then, we discuss iterative solution of the reduced BP equations, that is construction  of the MPD algorithm itself, in Section \ref{sec:MP-r-BP}. 

\vspace{-0.3cm}
\subsection{From Integral to Algebraic BP Equations}
\label{sec:functional-algebraic}\vspace{-0.1cm}

In order to reduce complexity of Eqs.~(\ref{eq:BP_factor_1_NN},\ref{eq:BP_factor_2_NN}), we simplify them taking advantage of the fact that the number of samples, $S$, is sufficiently large.  In the statistical mechanic jargon, we consider the \textbf{thermodynamic limit}, $S\to\infty$. It is important that we send $S$ to $\infty$ first,  i.e. before taking the limit of zero temperature (regularization parameter), $\beta\to\infty$, and also before considering the $P\to\infty$ limit of the large number of the functional parameters. The assumption, $S\gg \beta$ allows to guarantee universality of the solution,  and therefore its robustness to small changes in the samples, $\mathbb{IO}^{(S)}$.  Specifically, keeping $\beta$ large but finite accounts  for "entropic" configurations which are close to the ground state. On the other hand setting,  $S\gg P$,  aims at keeping the number of parameters (much) smaller than the number of samples to avoid over-fitting. Observe that, for all $(s\to p)\in \mathcal{E}$, the message $m_{p \to s}$ in Eq.~(\ref{eq:BP_factor_2_NN}) is a product  of $S-1$ positive functions. Therefore, keeping  $\beta$ finite and analyzing the case of sufficiently large $S$, i.e. $S\gg \beta$, we can utilize the Large Deviation (LD) approach:
\begin{equation}\label{eq:n-LD}
m_{p\to s}(\theta_p)\sim 
\exp\left(-S\beta h_{p \to s}(\theta_p)\right),
\end{equation}
where $h_{p\to s}(\theta_p)$ is a rate (Cram\'er) function,  depending (in the limit) only on a single parameter $\theta_p$. % and it is independent of $N$ (in the limit). 
The asymptotic LD structure of $h_{p\to s}(\theta_p)$ (\ref{eq:n-LD}) justifies the use of the Laplace method to approximate integration on the right hand side of Eq.~(\ref{eq:BP_factor_1_NN}). This means that, for each factor $s\in \mathcal{F}$ and vertex $p\in\mathcal{V}$, we approximate all the functions $m_{p'\to s}(\theta_p')$, with $p'\in \mathcal{V}/\{p\}$, appearing in the integral \eqref{eq:BP_factor_1_NN}, as a Gaussian distribution, centered around their maximum
\begin{align}
 m_{p'\to s}(\theta_p')  &=\sqrt{\frac{S\beta h''_{p\to s}}{2\pi }}e^{-\frac{S}{2} \beta h''_{p\to s}(\theta_p-\theta_{p\to s}^*)^2},\\
\label{eq:Gauss_rep1a}
\theta^*_{p'\to s}  & =\underset{\theta_p'\in \mathbb{R}}{ \text{argmin}}\, h_{p'\to s}(\theta_p')
\end{align}
and $h''_{p\to s}$ is the second derivative of $h_{p\to s}(\theta_p')$ with respect $\theta_p'$, computed at $\theta^*_{p'\to s}$. Continue to follow the  Laplace principe, for all $s\in{\cal F}$ and $p\in{\cal V}$, we derive
\begin{align}
\label{eq:BP_factor_1_NN_bis}
    \quad u_{s\to p}(\theta_p) & =-\lim_{S\to \infty}\frac{1}{\beta S}\log\left(m_{s\to p}(\theta_p)\right)\\ \nonumber & \sim l_s(\theta_p,\theta^*_{\partial p\to s})+O(1/S)\,,\\  \nonumber\theta^*_{\partial p\to s} & \doteq (\theta^*_{p'\to s}|p'\in {\cal V}\setminus \{p\}),\\ \nonumber l_s(\bm{\theta})  & \doteq l(\bm{x}^{(s)},\bm{y}^{(s)},\bm{\theta}),\\  \nonumber l_s(\theta_p,\theta^*_{\partial p\to s}) & =l(\bm{x}^{(s)},\bm{y}^{(s)},\bm{\theta})\big|_{\theta_{\partial p}=\theta^*_{\partial p\to s}}.
\end{align}
Rewriting the BP equation for the logarithms of the messages, and substituting \eqref{eq:Gauss_rep1a} and \eqref{eq:BP_factor_1_NN_bis} in \eqref{eq:BP_factor_1_NN} and \eqref{eq:BP_factor_2_NN}, we arrive at the following closed system of the asymptotic equations: $\forall s\in \mathcal{F},\ \forall p\in \mathcal{V}$,
\begin{align}
\label{eq:u_BP}
 \ u_{s\to p}(\theta_p) &=l_{s}( \theta_p,\theta^*_{\partial p\to s}),\\
\label{eq:h_BP}
h_{p \to s} (\theta_p) &=\frac{1}{S}\sum_{s'\in \mathcal{F}/\{s\}}u_{s'\to p}(\theta_p).\\
\label{eq:global_min_BP}
\theta^*_{ p\to s} &=\underset{\theta_p'\in \mathbb{R}}{ \text{argmin}}\, h_{p'\to s}(\theta_p')\,.
\end{align}
We will simplify Eqs.~(\ref{eq:u_BP},\ref{eq:h_BP},\ref{eq:global_min_BP}), transitioning to the objects averaged over the samples, $\forall p\in\mathcal{V}$: 
$\theta^*_{p} =\underset{\theta'_p\in \mathbb{R}}{ \text{argmin}}\, h_{p}(\theta_p'),\ 
    h_{p} (\theta_p) =\frac{1}{S}\sum_{s'\in \mathcal{F}}u_{s'\to p}(\theta_p)$.
This transformation is justified because according to \eqref{eq:h_BP},
$h_{p\to s} (\theta_p)-h_{p}(\theta_p)=\frac{u_{s\to p}(\theta_p)}{S}=O(1/S)$, and 
$\theta^*_{p\to s}=\theta^*_p+O(1/S)$, therefore resulting in replacement 
of the system of Eqs.~(\ref{eq:u_BP}-\ref{eq:global_min_BP}) over the $S\times P$ parameters, $\{\theta_{p\to s}| (p,s)\in \mathcal{E}\,\}$, by the following system of the \textbf{reduced Belief Propagation (r-BP)} equations over the $P$ parameters $\{\theta_{p\to s}| (p,s)\in \mathcal{E}\,\}$ 
\begin{align}
\label{eq:u_bp}
 \ u_{s\to p}(\theta_p) &=l_{s}( \theta_p,\theta^*_{\partial p}),\\
\label{eq:h_bp}
h_{p} (\theta_p) &=\frac{1}{S}\sum_{s'\in \mathcal{F}/\{s\}}u_{s'\to p}(\theta_p).\\
\label{eq:global_min_bp}
\theta^*_{ p} &=\underset{\theta_p'\in \mathbb{R}}{ \text{argmin}}\, h_{p}(\theta'_p)\,.
\end{align}
It is important to emphasize that the optimizations on the right hand side of the r-BP Eqs. \eqref{eq:global_min_bp} is one-dimensional. However, the optimization functions are non-convex, and moreover potentially rugged,  i.e. with multiple minima and maxima. Furthermore, functions on the right hand side of the Eqs.~(\ref{eq:global_min_bp}) are dependent on each other due to the relations (\ref{eq:u_bp},\ref{eq:h_bp}). Looking for ways to solve the system of Eqs.~(\ref{eq:global_min_bp}) we need to make sure that (a) each of the $S\times P$ optimization equations are satisfied individually, and (b) the results are syncronized.

\vspace{-0.2cm}
\subsection{Iterative solution of the r-BP Equations}\label{sec:MP-r-BP}
\vspace{-0.2cm}

We solve the r-BP equations Eqs.~(\ref{eq:u_bp},\ref{eq:h_bp},\ref{eq:global_min_bp}) iteratively, exploiting the directed nature of the messages, $u_{s\to p}$, 
%and $h_{p}$ from the factors to the vertices, and vice-versa, of 
over the graph $\mathcal{G}$. This approach results in a message-passing algorithm that makes a gradual descent in the loss-function landscape toward progressively more optimal configurations of the parameters. We initialize the parameters, $\theta^{*}_{p}$, for all $p\in \mathcal{V}$, at random. In the enabling case of the NN we choose to work with the zero mean Gaussian distribution %with the variance 
described in \cite{Kaiming}. Our Message Passing Descent (MPD) iterative implementation of the r-BP is as follows. At each iteration $t\in [T]$, we select a parameter $p_t\in \mathcal{V}$ at random. Then, we update the value of $\theta^{(t)}_{p_t}$, utilizing Eqs.~(\ref{eq:u_bp},\ref{eq:h_bp},\ref{eq:global_min_bp}), where $\theta^{*}_{\partial p_t}$, on the right-hand side is replaced with the values from the previous iteration:
\begin{align}
\label{eq:u_t}
 \ u^{(t)}_{s\to p_t}(\theta_{p_t}) & =l_{s_t}( \theta_{p_t},\theta^{(t-1)}_{\partial {p_{t-1}}}),\quad \forall s\in \mathcal{F}\\
\label{eq:h_t}
h_{p_{t-1}} (\theta_{p_t}) & =\frac{1}{S}\sum_{s'\in \mathcal{F}/\{s_t\}}u_{s'\to p_t}(\theta_p),\\
\label{eq:global_min_t}
\theta^{(t)}_{ p_t} & =\underset{\theta_{p_t}'\in \mathbb{R}}{ \text{argmin}}\, h_{p_t}(\theta'_{p_t})\,.
\end{align}
%We call the iterative algorithm the \textbf{Message Passing Descent (MPD)}.  

As already mentioned, the MPD implements a global optimization. This implies that, in contrast with the GD, the local minima of the MPD are not fixed points of the iterative process. As a consequence, the rugged/glassy landscape of the LF is not a handicap to the MPD, i.e. MPD iterations will not  be stuck in local minima, since the non-locality of the global optimization, at each iteration step $t$, allows to escape from the basin of a local minimum and explore a wider region of the configuration space. After a certain number of iterations $T\in \mathbb{N}$ the value of the LF does not decrease appreciably any further. However, even in this regime, the configuration of the parameters $\bm{\theta}^{(t)}$ may be still far from convergence, i.e., for $t>T$, the updated value $\theta^{(t)}_{p_t}$ may still deviates considerably from its previous value $\theta^{(t-1)}_{p_t}$. In particular, the $\bm{\theta}^{(t)}$ may not converge at all and the MPD algorithm continues indefinitely to jumps amongst different configurations with the same value of the LF. In any case, since the value of LF is the same, the resulting MF $g[\bm{\theta}^{(t)}]$, with $t>T$, provide the same degree of approximation of the data-set; so we can stop the iteration regardless the convergence in configuration. The  final configuration, $\bm{\theta}^T=(\theta^{T}_{p}|p\in \mathcal{V})$, is the MPD algorithm output. Even though transformation from the BP Eqs.~(\ref{eq:BP_factor_1_NN},\ref{eq:BP_factor_2_NN})  to the iterative MPD Eqs.~(\ref{eq:u_t},\ref{eq:h_t},\ref{eq:global_min_t}) is a major simplification (as the multi-dimensional integration is replaced by a one-dimensional optimization), the iterative Eqs.~(\ref{eq:global_min_bp},\ref{eq:u_bp},\ref{eq:h_bp}) still constitute a challenge for efficient implementation, because the remaining are functional, specifically requiring to extract at each step of the iterative procedure a function of a single parameter from multivariate functions of all the parameters.  Resolving of the remaining computational challenges are discussed in the following two Sections.

\vspace{-0.2cm}
\section{MPD over PWP Functions}
\label{sec:piece-wise}
\vspace{-0.2cm}

This Section is split in three Subsections. Section \ref{sec:PWP-motivation} gives some additional rationale for using the Piece-Wise-Polynomial (PWP) approximation.  We introduce PWP notations for the functions entering the iterative MPD Eqs.~(\ref{eq:u_t},\ref{eq:h_t},\ref{eq:global_min_t}) in Section \ref{sec:PWP-notations}. Then, we recast the MPD equations over  the PWP functions in  the form of the pseudo-algorithm in Section \ref{sec:algorithms} (see also the Appendix \eqref{appendix}).

\vspace{-0.2cm}
\subsection{PWP Representation: Motivation}\label{sec:PWP-motivation}
\vspace{-0.2cm}

We focus on the case when messages are all PWP in the parameters over a finite number of sub-domains. This PWP assumption allows us to reduce the iterative MPD Eqs.~(\ref{eq:u_t},\ref{eq:h_t},\ref{eq:global_min_t}) to a set of algebraic relations over the set of separation points of the sub-domains (mesh) and the polynomial coefficients, corresponding to each sub-domain. Moreover, if $h^{(t)}_{p_t\to s_t}$ is piece-wise quadratic the stationary points within each sub-domain can be computed exactly, by solving respective quadratic equations. The PWP assumption is exact in the case where the model function, $g[\bm{\theta}]$, in \eqref{eq:g-model} is a Piece-Wise-Linear (PWL) with respect the parameter and the LL \eqref{eq:LL} is at most quadratic, as in Eq.~(\ref{eq:LL-l2}). Otherwise, we can consider a PWP interpolation of the MF and the LF. 

%Our enabling example of the PWL model function, $g[\bm{\theta}]$, used  in the experiments of Section \ref{sec:experiments} will correspond to the case of the Feed-Forward Neural Network (FF-NN) with the Leaky  Rectified Linear Unit (Leaky-ReLU) activation functions.

\vspace{-0.2cm}
\subsection{PWP Representation: Notations}\label{sec:PWP-notations}
\vspace{-0.2cm}

Assuming that $l_{s_t}({\bm \theta}_p)$, entering Eq.~(\ref{eq:u_t}), is a multi-variate PWP function of the vector of parameters, ${\bm\theta}_p$, with a finite number of sub-domains, we arrive at the one dimensional function, $u_{s\to p}(\theta_p)$, of a particular parameter, $\theta_p$, which is also PWP over a finite number of sub-domains. (Here and below we use shortcut notations dropping, $t$, index for all relevant characteristics changing in iterations. We assume that the value is updated to the current $t$-dependent status.) Consider the following parametrization of the PWP, $u_{s\to p}(\theta_p)$:
\begin{gather}
\label{eq:u_poly}
u_{s\to p}(\theta_p)=\sum^{R}_{r=1} \chi_{[\vartheta^{(r)}_{s\to p},\vartheta^{(r+1)}_{s\to p}]}\left(\sum^Q_{q=0}c^{(r,q)}_{s\to p} \theta_p^q\right)\,,
\end{gather}
where we brake the domain of $\theta_p$ into $R+1$ sub-domains each modeled by a polynomial of the degree $Q$; boundaries of the sub-domains are described via the ordered list of real numbers, $(\vartheta^{(r)}_{s\to p}|r \in [R+1])$, where we also set, $\vartheta^{(1)}_{s\to p}=-\infty$ and $\vartheta^{(R+1)}_{s\to p}=\infty$; $\chi_{[\vartheta^{(r)}_{s\to p},\vartheta^{(r+1)}_{s\to p}]}$, denotes the indicator function of the sub-domain $[\vartheta^{(r)}_{s\to p},\vartheta^{(r+1)}_{s\to p}]$, which is unity if the argument is within the sub-domain and zero otherwise. The PWP approximation (\ref{eq:u_poly}) of messages is described in terms of the following vectors of parameters
\begin{gather}\label{eq:s-to-p}
    \Theta_{s\to p}=\left(\vartheta^{(r)}_{s\to p}|r \in [R+1]\right)\,,\\C_{s\to p}=\left((c^{(r,0)}_{s\to p},\cdots,c^{(r,q)}_{s\to p})|r \in [R]\right).
\end{gather}
Furthermore, Eqs.~(\ref{eq:u_poly},\ref{eq:h_bp}) suggest that the $h_{p\to s}$-functions are PWP too
\begin{gather}
\label{eq:h_poly}
h_{p}(\theta_p)=\sum^{SR}_{r=1} \chi_{[\vartheta^{(r)}_{p},\vartheta^{(r+1)}_{p}]}\left(\sum^Q_{q=0}c^{(r,q)}_{p} \theta_p^q\right)\,,
\end{gather}
characterized by the following vector of parameters
\begin{gather}
  \label{eq:bbH}
    \Theta_{p}=\left(\vartheta^{(r)}_{p}|r \in [SR+1]\right)\,,\\
  \label{frakH}
     C_{p}=\left((c^{(r,0)}_{p},\cdots,c^{(r,q)}_{p})|r \in [SR]\right)\,.
\end{gather}
The PWP setup allows us to re-cast in the following the iterative MPD Eqs.~(\ref{eq:u_t},\ref{eq:h_t},\ref{eq:global_min_t}) in the form of an iterative algorithm which picks at each step a parameter $p\in \mathcal{V})$ at random and updates the value of the parameter $\theta_{p}$, within both the $\Theta_{p}$ and $C_{p}$ meshes. Notice, that even though our presentation of the algorithm is general, as applicable to any piece-wise-polynomial $u$- and $h$-functions entering Eqs.~(\ref{eq:u_poly},\ref{eq:h_poly}), actual implementation, and specifically computation of  $\Theta_{s\to p}$ and $C_{s\to p}$, depends explicitly on the choice of the MF \eqref{eq:g-model}  and of the loss-function \eqref{eq:loss_fun}. 

\vspace{-0.2cm}
\subsection{Message Passing Descent Algorithm: Pseudo-Code}
\label{sec:algorithms}\vspace{-0.2cm}

In this Subsection, we summarize and present the MPD Algorithm with the PSP messages (functions) formally and aiming to state it is the  "ready to implement" form. The MPD algorithm consists of the four sub-routines -- Algorithms 1-4 respectively (with Algorithms \eqref{alg:MPD-M} and \eqref{alg:GLOBAL-MIN} detailed in the Appendix).

At time $t=0$, we initialize parameters of the NN at random  in the Algorithm \ref{alg:initialization}.
\begin{algorithm}
\caption{Initialization}
\begin{algorithmic}[1] \label{alg:initialization}
\STATE  $\forall p\in \mathcal{V},\quad \theta_{p}\sim \mathcal{U}(-\sqrt{3\sigma},\sqrt{3\sigma})$.
\end{algorithmic}
\end{algorithm}
\noindent The distribution $\mathcal{U}(-\sqrt{3\sigma},\sqrt{3\sigma})$ is the uniform distribution, with the zero mean and the variance, $\sigma$. If the MF $g[\bm{\theta}]$ in \eqref{eq:g-model} is a NN, a convenient choice for $\sigma$ is provided by the Kaiming initialization \cite{Kaiming}.
%After initialization, we run the MPD Algorithm for $T$ iterations.
We brake the remainder of the MPD algorithm in three parts, constituting the main iterative step of MPD in Algorithm \ref{alg:MPD} and its two sub-routines presented in the Appendix \eqref{appendix}: Algorithm 3 -- subroutine MPD-SUM \eqref{alg:MPD-M}-- 
computing $\Theta_{p\to s}$ and $C_{p\to s}$, and Algorithm 4 -- subroutine GLOBAL-MIN \eqref{alg:GLOBAL-MIN}-- % 
executing Global one-dimensional optimization. 
\begin{algorithm}
\caption{MPD iteration:}
\textbf{Input:} Initial configuration $\bm{\theta}_p$, data-set $\mathbb{IO}^{(S)}$,\\
\textbf{Output:} $\bm{\theta}^*_p$.
\begin{algorithmic}[1] \label{alg:MPD}
\FOR{$t \in [T]$}
        \STATE Pick a vertex $p\in \mathcal{V}$: $p\sim U(\mathcal{V})$.
       \STATE 
       %$\forall s' \in \mathcal{F}/\{s\}$, compute $\Theta_{s'\to p}$ and $C_{s'\to p}$
       $\forall s \in \mathcal{F}$, compute $\Theta_{ s\to p}$ and $C_{s\to p}$\\from $\bm{\theta}^{(t-1)}_{\partial {p_t}}$ and the IO sample $(\bm{x}^{(s)},\bm{y}^{(s)})$.
        \STATE Compute $\Theta_{p}$ and $C_{p}$ from $\{(\Theta_{s\to p},C_{s\to p})|\,s\in\mathcal{F}\}$. \\ \rightline{\COMMENT{Use subroutine MPD-SUM.}}
        \STATE  Update $\theta_{p}$ from $\Theta_{p}$ and $C_{p}$: \\ \rightline{$\theta_{p}\leftarrow \text{argmin}_{\widehat{\theta}_p}(h_{p}(\, \widehat{\theta}_p)\,).$} \rightline{\COMMENT{Use subroutine GLOBAL-MIN.}}\rightline{\COMMENT{Do not update other parameters.}}
%       \STATE  $\forall (p'\to s') \neq (p\to s),\quad  \theta_{p'\to s',t} =\theta_{p'\to s',t-1},$  \\ \rightline{\COMMENT{Do not update other parameters.}}
\ENDFOR
        \RETURN Result:  $\theta^*_p =\theta_p,\quad \forall p\in \mathcal{V}$
\end{algorithmic}
\end{algorithm}
Obtaining the mesh $\Theta_{p}$ and the  $C_{p}$ efficiently from the collections of meshes$\{\Theta_{s\to p}|s\in \mathcal{F}\,\}$ and lists of coefficients $\{C_{s\to p}|s\in \mathcal{F}\,\}$, according to Eqs.~(\ref{eq:h_bp}), is the most challenging part of the MPD Algorithm. (We omit the division by $S$, since this correction does not affect the result of the global minimization \eqref{eq:global_min_t}.)

Note that the mesh $\Theta_{p}$, is equal to the union of the $\{\Theta_{s\to p}|s\in \mathcal{F}\,\}$ meshes. This implies, in particular, that the left-most sub-domain of $h_{p}$, i.e. $[\vartheta^{(0)}_{p\to s},\vartheta^{(1)}_{p\to s}]=[-\infty,\vartheta^{(1)}_{p\to s}]$, is a subset of all the left-most sub-domains of the messages $u_{s\to p}$, with $s\in \mathcal{F}$:
\begin{equation}
    c^{(0,q)}_{p}=\sum_{s\in \mathcal{F}} c^{(0,q)}_{s\to p}\,.
\end{equation}
In each of the sub-domains a mesh point, $\vartheta^{(r)}_{p}$, of $h_{p}$ coincides with a mesh point of $u_{s\to p}$, for some $s \in \mathcal{F}$, unless a degeneracy. In other words, for each mesh point $\vartheta^{(r)}_{p}$ of $h_{p}$, there exists just one factor $s=s(r)\in \mathcal{F}$ and a number $r'=r'(r)\in [R]$ such that $\vartheta^{(r'(r)}_{s(r) \to p}=\vartheta^{(r)}_{p}$. The relations suggest that the coefficients $c^{(r,q)}_{p}$, for $r\in[R]$ and $q\in [Q]$, satisfy
\begin{gather*}
 c^{(r,q)}_{p}=c^{(r-1,q)}_{p}+c^{(r'(r),q)}_{s(r)\to p}-c^{(r'(r)-1,q)}_{s'(r)\to p}.
\end{gather*}

%The Algorithm \ref{alg:MPD} performs a batch minimization of the LF. 
Note that, at each step, the MPD algorithm updates just one parameter. As a consequence, the basic version of the MPD algorithm requires, in order to update all the parameters, at least  $O(P)$ iterations. Then, the entire computational cost of the basic MPD Algorithm \ref{alg:MPD} becomes at least, $O(Q \times R\times S\times \log(S) \times P)$. However, and as argued below, the computational cost can be reduced significantly if we utilize a mini-batch version of the MPD.

The mini-batch implementation of the MPD algorithm consists in selecting, at each step of the Algorithm \ref{alg:MPD}, a random sub-set of the data-set (so-called mini-batch), of size $S'<S$  (called the mini-batch size). Otherwise, we simply follow all the sub-steps of the Algorithm \ref{alg:MPD}, e.g. computing all the required meshes and the coefficients for the mini-batch (and not for the entire data-set). Formally, the mini-batch version of the MPD Algorithm \ref{alg:MPD} consists in replacing line $3$ with\\
\hspace*{0.2cm} 3'.1: Select a random subset $\mathcal{F}'_t\subset \mathcal{F}$ of size $S'$\\
\hspace*{0.2cm} 3'.2: $\forall s' \in \mathcal{F}'_t$: Compute $\Theta_{ s'\to p},\ C_{s'\to p}$ from\\
\hspace*{0.9cm} $\bm{\theta}^{(t-1)}_{\partial {p_t}}$ and the IO sample $(\bm{x}^{(s')},\bm{y}^{(s')})$\\
and then replace line $4$ with\\
\hspace*{0.2cm} 4': Compute $\Theta_{p},\ C_{p}$ from $\{(\Theta_{s'\to p},C_{s'\to p})|\,s'\in\mathcal{F}'\}$.\\
Computational cost of these corrections is $O(Q \times R\times S'\times \log(S'))$ and then the entire cost of the mini-batch MPD becomes $O(Q \times R\times S'\times \log(S') \times P)$. We choose the subsets $\mathcal{F}'_t$, for any integer number $t'$:
\begin{equation}
\label{eq:minibath}
  \mathcal{F}=  \bigcup^{(t'+1)\left\lceil{\frac{S}{S'}}\right\rceil-1}_{t=t'\left\lceil{\frac{S}{S'}}\right\rceil}\mathcal{F}'_t
\end{equation}
In such a way, we periodically select all the samples. We will refer to the MPD step over a mini-batch $\mathcal{F}_t$ as the mini-batch step, and we also call the group of the $\lceil{S/S'}\rceil$ consecutive steps in the union \eqref{eq:minibath} the batch step. Notice that the mini-batch trick improves performance only if the mini-batch size, $S'$, is chosen carefully. Indeed, decrease of the LF observed at each step of the basic MPD algorithm is linked to the fact that, according to Eqs.~ (\ref{eq:u_bp},\ref{eq:h_bp}) the message function $h_p(\theta_p)$ is the LF considered as a function of a single parameter $\theta_p$. However, when we replace the batch $\mathcal{F}$ with a mini-batch, this is no longer true, i.e.
$h^{\text{mini-batch}}_p(\theta_p) =\frac{\sum_{s'\in \mathcal{F}'} l_{s'}(\theta_p,\bm{\theta}^*_{\partial p})}{S'}\neq \frac{\sum_{s\in \mathcal{F}} l_s(\theta_p,\bm{\theta}^*_{\partial p})}{S} =h^{\text{batch}}_p(\theta_p)$,
and one arrives at the desirable asymptotic, $h^{\text{mini-batch}}\to h^{\text{mini-batch}}$, only in the $S,S'\to \infty$ limit. In the next Section, devoted to Numerical Experiments, we show how a practical compromise is achieved at, $1\ll S' \ll S$.

\vspace{-0.2cm}
\section{Experiments}\label{sec:experiments}\vspace{-0.2cm}

The MPD algorithm was introduced and discussed in details in Section \ref{sec:piece-wise} in the general case of the PWP Model Function, $g$, entering the parameter fitting formulation via Eq.~(\ref{eq:LL}) and defined in Eq. \eqref{eq:g-model}. In this Section devoted to experimental testing and validation of the MPD algorithm we choose to work with the special form of the MF -- a FFNN with a PWL activation function. %We also limit our experiments to the cases with FFNN with one and two hidden layers. 
Our choice of the loss function is the $L_2$, described in Eq.~(\ref{eq:LL-l2}).  For a FFNN with one layer, we are able to compute the exact list of coefficients and the mesh analytically, and thus efficiently. Therefore, this Section, dedicated to the experiments in the aforementioned setting is split in two Subsection. We describe the PWL structure of the FFNN in Section \ref{sec:PWL-FFNN}. Our numerical experiments, describing a mini-batch version of the Algorithm \ref{alg:MPD} are presented in Section \ref{sec:experiments}.

\subsection{PWL structure of FFNN}\label{sec:PWL-FFNN}

We substitute the general MF, $g$ (defined in Eq.~(\ref{eq:LL}) as a function that applies to the vector, $\bm{x}\in \mathcal{X}$)  by a FFNN, $\textbf{NN}:\mathcal{X}\longrightarrow \mathcal{Y}$, which is a nested composition of functions represented by alternating linear and non-linear layers (of functions). Let $N$ be the depth of the FFNN, where the input layer and the output layer are labeled by $0$ and $N$ respectively. Let $f:\mathbb{R}\longrightarrow\mathbb{R}$ be the activation function (AF), which we choose to be (without loss of generality) a PWL function; let $d_n$ be the width of the $n-$th layers, with $0\leq n\leq N$. Within this notation, each input vector $\bm{x}^{(s)}$ has the dimension $d_{\text{in}}=d_0$ and each output vector $\bm{y}^{(s)}$ has the dimension $d_{\text{out}}=d_{N}$. The FFNN activation function, $g(\bm{x}^{(s)})\to\textbf{NN}(\bm{x}^{(s)})$, can be formally stated in terms of the following recurrence 
\begin{align}
     \label{eq:H_0}
     \bm{z}^{(s)}_{0} & \doteq\bm{x}^{(s)}\in \mathbb{R}^{d_0},\\
     \label{eq:H_n_lin}
     \widetilde{\bm{z}}^{(s)}_{n} & \doteq\widehat{w}_{n}\bm{z}^{(s)}_{n-1}+\bm{b}_{n}\in \mathbb{R}^{d_n}, \quad (0<n<N)\\
     \label{eq:H_n_nlin}
     \bm{z}^{(s)}_{n} & \doteq f(\widetilde{\bm{z}}^{(s)}_{n})\in \mathbb{R}^{d_n}, \quad (0<n<N),
\end{align}
where the AF, $f$, acts component-wise; $\widehat{w}_n$ is a $d_{n-1}\times d_n$ matrix with components $(w_{nd'd''},\,d'\in [d_n],\,d''\in[d_{n-1}])$ (the weights); and $\bm{b}_n$ is a $d_n$-component vector, $(b_{n1},\cdots,b_{nd_{n}})$, (of the biases). 
Let us use $\mathcal{W}$ and $\mathcal{B}$ for  the lists of all the weights and biases respectively, and $\mathcal{W}_n$ and $\mathcal{B}_n$ for the lists of weights and biases of the layer $n\in [N]$. %Let also set $\mathcal{W}_{nd'}\doteq(w_{nd'd''},\,d''\in[d_{n-1}])$ and $\widetilde{z}^{(s)}_{nd'}$ (resp. $z^{(s)}_{nd'}$), with $d'\in [d_{n'}]$, for the components (nodes) of $\widetilde{\bm{z}}^{(s)}_{n}$ (resp. $\bm{z}^{(s)}_{n}$). 
Then the FFNN is 
\begin{gather}
    \label{eq:NN}
   \!\!\!\textbf{NN}\left[\bm{\theta}=(\mathcal{W},\mathcal{B})\right](\bm{x}^{(s)})\doteq\widetilde{\bm{z}}^{(s)}_{N}\doteq\widehat{w}_{N}\bm{z}^{(s)}_{N-1}+\bm{b}_{N}.
\end{gather}
Since the space of the PWL function is closed over linear transformations, as in Eq.~(\ref{eq:H_n_lin}), and over the composition transformations, as in Eq.~(\ref{eq:H_n_nlin}) with a PWL function $f$, the combination of the two operations applied sequentially,  as in Eqs.~(\ref{eq:H_0},\ref{eq:H_n_lin},\ref{eq:H_n_nlin},\ref{eq:NN}), results in the PWL model function and our task becomes to compute it efficiently. Let us recall that due to the factorized nature of the MDP construction, presented above, we need to have an efficient way of evaluating  the $u$-message functions, defined in Eq.~(\ref{eq:u_bp}), as a function of a particular parameter, $\theta_p\in \mathcal{W}_n\times \mathcal{B}_n$,  when all other parameters are fixed (to their current values in the process of the MDP execution) and do it for a particular sample, $s$. In the case of the FF-NN (\ref{eq:NN}) and of the $l_2$ loss function (\ref{eq:LL-l2}), the respective expression becomes
\begin{gather}\label{eq:u_s-p-lin}
    u_{s\to p}(\theta_p)=\left|{\bm y}^{(s)}-\tilde{\bm z}_{N,d}^{(s)}(\theta_p)\right|^2,
\end{gather}
where the dependence on $\theta_p$, enters implicitly via the recurrence according to Eqs.~(\ref{eq:H_0},\ref{eq:H_n_lin},\ref{eq:H_n_nlin},\ref{eq:NN}). The structure of the NN, just presented,  makes it obvious that the $z$ values at the hidden nodes (current readings of the neurons), $\widetilde{z}^{(s)}_{n'd'}$ and $z^{(s)}_{n'd'}$, with $n'>n$ and $d'\in [d_{n'}]$, are PWL functions of $\theta_p$, while for $n'<n$, they   are independent of $\theta_p$. 
In order to represent $\widetilde{\bm{z}}_N^{(s)}$ as a PWL function of $\theta_p$ in the standard form, i.e. compute its coefficients and mesh (see e.g. the right-hand-side of Eq.~(\ref{eq:u_poly}) with $Q=1$), we need  to analyze the linear combination \eqref{eq:H_n_lin} and the composition \eqref{eq:H_n_nlin} of the PWL functions acting on the values of $z$ from the preceding layer. We start the analysis from the linear combination and split it into three parts: multiplication on the weights, sum over the rows and addition of the biases. We act by induction: set $n'>n$ and assume that the mesh and the coefficients are already constructed for the layer $n'-1$. Let us create the matrix of the PWP functions, $(w_{n'd'd''} z^{(s)}_{n'-1,d'}[\theta_p],\,d'\in [d_n],\,d''\in[d_{n-1}])$, where obviously, $w_{n'd'd''} z^{(s)}_{n'-1,d'}[\theta_p]$, has the same mesh as $z_{n'-1,d'}^{(s)}[\theta_p]$ and the coefficients extracted from, $z_{n'-1,d''}^{(s)}[\theta_p]$, multiplied by $w_{n'd'd''}$. Next, we compute the mesh and the coefficients of the sum over the rows $\sum_{d''\in [d_{n'-1}]}w_{n'd'd''}z^{(s)}_{n',d''}[\theta_p]$, by applying the Algorithm \eqref{alg:MPD-M}  in the Appendix %
to the vector of the PWL functions, $(w_{n'd'd''} z^{(s)}_{n'-1,d''}[\theta_p],\ d''\in [d_{n'-1}])$. We finally add the bias $b_{d'd''}$ to the $0$-degree coefficients, thus arriving  at the  respective expression for the mesh and for the coefficients of $\widetilde{z}^{(s)}_{n',d'}[\theta_p]$. Let us now consider the respective functional composition \eqref{eq:H_n_nlin}. Boundaries of the linear sub-domains of the composition are found by solving 
\begin{gather}\label{eq:mesh_compose}
\forall n'>n,\ \forall d'\in [d_{n'}]:\ \widetilde{z}^{(s)}_{n',d'}[\theta_p]=0.
\end{gather}
The sorted union of the solutions of \eqref{eq:mesh_compose} and the mesh of $\widetilde{z}^{(s)}_{n',d'}[\theta_p]$ produce the desired mesh for the PWL, $z^{(s)}_{n',d'}[\theta_p]$. The exact coefficients are found computing, $z^{(s)}_{n',d'}[\theta_p]$, at the points of the mesh and connecting any nearest neighbor pairs of points via a linear function. Repeating the three step process inductively , till $n'=N-1$, results in the desired construction of the standard PWL representation for $\tilde{\bm z}_{N,d}^{(s)}(\theta_p)$ in Eq.~(\ref{eq:u_s-p-lin}). Notice that Eq.~(\ref{eq:mesh_compose}) may have a separate solution in each sub-domain of $\widetilde{z}^{(s)}_{n',d'}[\theta_p]$. This means that the composition may double the number of points of the mesh. Accounting for the fact that the linear combination of PWL is described by a union of meshes we estimate that the number of sub-domains for each, $\bm{z}^{(s)}_n$, is $R_{n}\sim 2d_{n-1}R_{n-1}$, i.e. it grows exponentially with $N$, making exact computations prohibitively expansive for the case of a deep NN, where $N\gg 1$. We postpone discussion of an approximate evaluation of the PWL FWNN with large number of layers to future publications, and describe in the next Subsection our experiments with one layer,  where FWNN and its derivatives are evaluated efficiently and analytically, according to the construction which we have just completed presenting in this Subsection.  
\vspace{-0.2cm}
\subsection{Numerical Experiments}\label{sec:experiments}\vspace{-0.2cm}
Here we report results of our numerical experiment comparing MPD training with the state-of-the-art GD training on the example of the FFNN with one hidden layer. The experiments were implemented in PyTorch \cite{pyThorch1,pyThorch2}. The activation functions (in the non-linear sub-layer) are leaky versions of the “hard tanh” function,
\begin{equation}
f(x)=\begin{cases}
\alpha x+ \text{sign}(x)(1-\alpha)\,,\quad |x|>1,\\
x\,,\quad |x|\leq1\,
\end{cases}
\end{equation}
with $\alpha=0.01$. (The leaky component with small slope at $|x|>1$ is beneficial for convergence of all the training algorithms considered.) We work with the mean square error ($L_2$) loss-function (correspondent to the standard regression setting). We have selected for this experiment a subset of the 3D Road Network (of North Jutland, Denmark) data-set \cite{data_set}, from the UCI Machine Learning repository \cite{UCI}, containing $50929$ samples  over the region with the total area of, $10\times 7\,km^2$. This particular subregion is chosen for its roughness (containing a number of irregular patterns) thus (conjectured) leading to a sufficiently rough loss-function parameter landscape containing multiple minima, saddle-points and maxima. Each sample in this data set is represented via the input --- two dimensional vector ($d_0=2$) of the longitude and latitude of the street crossings within a two-dimensional network of roads --- and the output -- one dimensional vector  representing respective vertical elevations ($d_{\text{out}}=1$). 
%The complete data-set covers a region of $185\times 135 km^2$ in North Jutland, Denmark. This region includes many uncorrelated  local patterns (hills and dells) thus making the regression problem challenging. %We consider smaller area of the whole region.
%We also substract mean and normalize the data,  so 
The samples are modified to guarantee zero mean and unit variance. 
\begin{comment}
the original inputs and outputs $\{(\bm{x}^{(s)},\bm{y}^{(s)})|\,s\in[S]\}$, %obtaining the normalized data-set $\{(\widetilde{\bm{x}}^{(s)},\widetilde{\bm{x}}^{(s)})|\,s\in[S]\}$ in such a way: 
according to
\begin{gather}
\widetilde{x}^{(s)}_d=\frac{x^{(s)}_d-\overline{x^{(s')}_d}}{\sqrt{\overline{(x^{(s')}_d)^2}-\overline{x^{(s')}_d}^2}}\,,\quad \text{for} \, d\in[d_0],s\in[S]\\ 
\widetilde{y}^{(s)}_d=\frac{y^{(s)}_d-\overline{y^{(s')}_d}}{\sqrt{\overline{(y^{(s')}_d)^2}-\overline{y^{(s')}_d}^2}}\,,\quad \text{for} \, d\in[d_0],s\in[S]\,,
\end{gather}
where the symbol $\overline{\,\cdot\,}$ denotes averaging over the samples $s'\in[S]$. 
We train the FFNN on a (relatively small) subset of the original data-set
\begin{equation}
\mathbb{IO}=\{(\widetilde{\bm{x}}^{(s)},\widetilde{\bm{y}}^{(s)})|\,s\in[S],\quad  \widetilde{\bm{x}}^{(s)}[-0.1,0.1]^2\, \}
\end{equation}
\end{comment}
We use $80\%$ of samples for training and the remaining $20\%$ for validation.
%We also use a relatively small subset of the original data set containing
%The resulting data-set $\mathbb{IO}$ contains 
%$50929$ samples  over the region with the total area of, $10\times 7\,km^2$. This particular subregion is chosen for its roughness (containing a number of irregular patterns) thus (conjectured) leading to a sufficiently rough loss-function parameter landscape containing multiple minima, saddle-points and maxima.
%This reduced data-set describes a region that is still rough with several pattern. We guess roughness on the data-set cause the loss-function to be a rough function with respect the parameters (presenting a lot of local-minima).

Width, i.e. number of neurons, in the (only) hidden layer of the NN is $d_1=500$, therefore resulting in the $S=2001$ parameters. We juxtapose the MPD algorithm to the Adam algorithm \cite{Adam} and to the Nesterov Accelerated Gradient (NAG) algorithm \cite{NAG}, which are arguably the two most popular state-of-the-art iterative algorithm of the Gradient Descent (GD) type utilized to train NNs. Comparing the three methods we use the mini-batch optimization for all. The two GD methods yields the best performance when the mini-batch size is $256$ and the learning rate is $10^{-3}$. The best performance of the MPD algorithm is attained when the min-batch is $2048$. After the initial $t=200$ iterations, we progressively increase the mini-batch size (following the standard guidance for reducing fluctuations in the LF). It is worth noting that, in general, the MPD algorithm requires a much larger mini-batch size then GD methods. This is due to the fact that random fluctuations, originating from the small mini-batch size, may help the GD algorithm to escape from the basin of a local minimum, and thus reach a better configuration. In contrast, the randomness may cause the global minimization sub-routine (see Algorithm \eqref{alg:GLOBAL-MIN} in the Appendix) %\eqref{alg:GLOBAL-MIN} 
to fail. In this regards,  mini-batching is used in the case of MPD solely for speeding it up.  

\begin{figure}[t]
   \centering
  \includegraphics[width=0.9\textwidth]{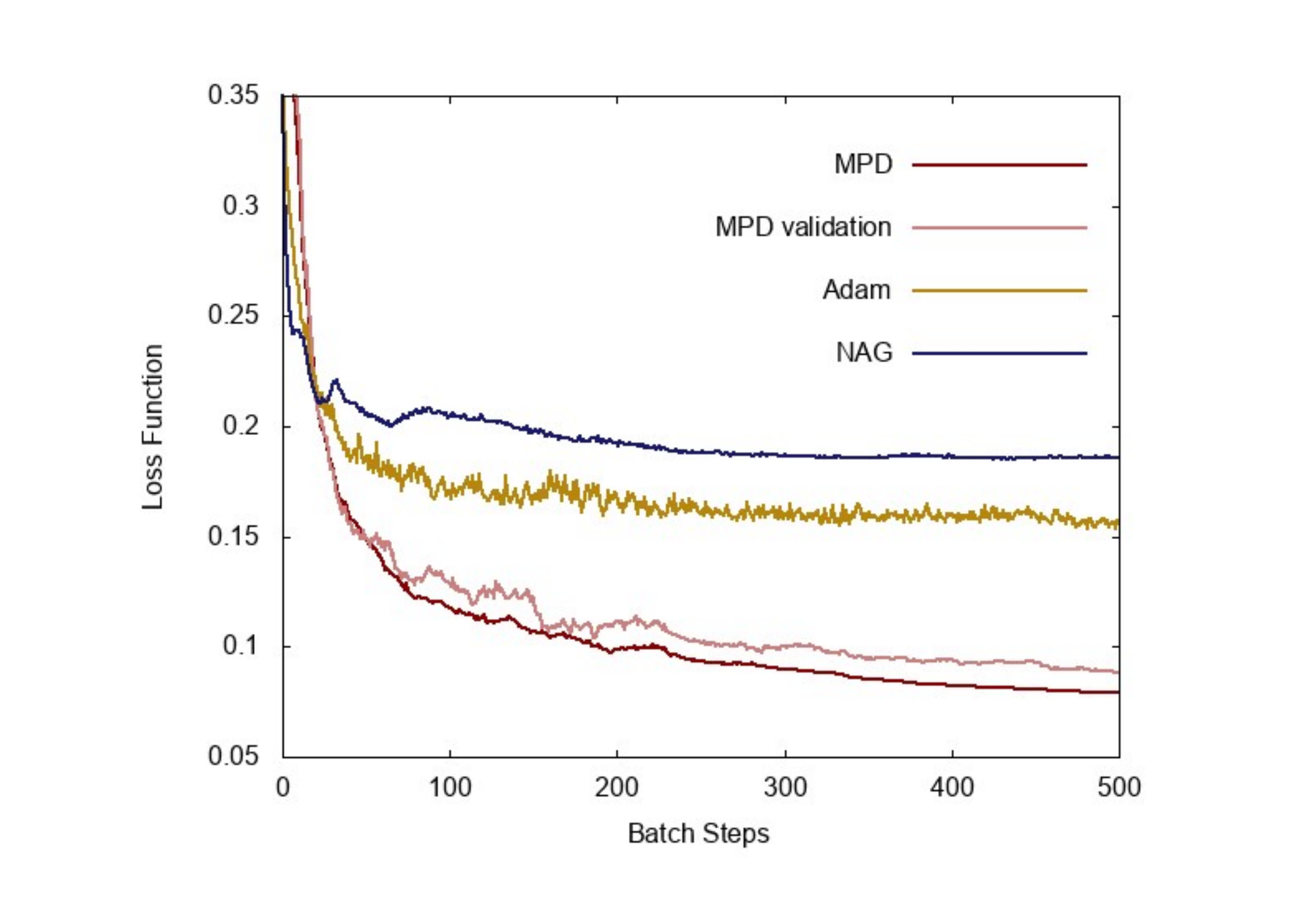}
\vspace{-1.2cm}
    \caption{Comparative analysis of MPD, Adam and NAG: dependence of the LF, shown on the y axis, on the number of iterations (each associated with a new batch), shown on the x axis. For MPD we show the training loss (MPD) and the validation loss (MPD validation).
    \label{fig:ploindt}}
\end{figure}

Results of our experiments are reported in Fig.~\ref{fig:ploindt}, showing dependence of the LF on the number of batch steps. In this (intentionally chosen) case of the data with sufficiently rugged parameter landscape we confirm our theoretical assertion:  the MPD algorithm,  making non-local steps, outperforms Adam and NAG, which both rely on the GD guidance -- thus local by-design. 

\section{Conclusions and Path Forward}\label{sec:conclusion}\vspace{-0.2cm}

We proposed a new optimization algorithm which applies to a wide class of optimization problems in ML,  e.g. training of NNs,  where we ought to fit data with a parameterized Model Function (MF) mapping input to the output. The most successful training algorithms in ML achieve the task of minimizing the loss function via a Gradient Descent. At each step, GD like algorithms update all the vector of parameters evaluating the gradient of the LF at the current, i.e. local, value of the vector via automatic differentiation. In a stark contrast with the state-of-the-art approach our newly suggested MPD algorithm advances resolving the LF optimization globally.  Moreover, this critical optimization step is done efficiently optimizing over each of the parameters sequentially, evaluating each of the single-parameter optimizations analytically. Ability to take global steps, thus avoiding trapping in undesirable local minima of the loss function landscape, and make the steps analytically are the two major advantages of the MPD algorithm over the GD like algorithms. These advantages of the MPD over the Adam algorithm (most practically successful algorithm of the GD type) are confirmed in experimental tests done on examples of training FFNN with one and two hidden layers.  

The MPD algorithm was derived in the manuscript in steps:  
\begin{itemize}
    \item First, we state the general data fitting problem as a Graphical Model;
    \item Second, we derive the Belief Propagation equations describing the process of training;
    \item Third, we simplify the BP equations,  which are functional, integral equations, to their reduced form, r-BP, taken advantage of the Large Deviation approach justified in the limit when the number of samples is significantly larger than then number of parameters (and both are large); 
    \item Fourth,  the r-BP equations are translated into iterative equations of the MPD algorithm requiring,  at each step of the iteration process, to evaluate one dimensional optimization over a single parameter,  fixing dependence of the MF on the other parameters fixed to their current value;
    \item Fifth, to make evaluation of the one dimensional optimization analytic we consider the case of the MF represented as a Piece-Wise-Polynomial (PWP) function,  including many NN examples,  e.g. described by FFNN, Convolutional Neural Networks, Graphical Neural Networks with PWP activation functions;
    \item Six, to show practical utility of the MPD algorithm, that is algorithmic advantage when evaluated against the state-of-the-art algorithm of the GD type, we experiment with examples of FFNN with one- and two- hidden-layers, LeakyReLU activation functions and $L_2$-loss function; These cases are special, as resulting in the PWL representation for the MF, then allowing analytic evaluation of the aforementioned single-parameter optimizations within the MPD algorithm.   
\end{itemize} 
It is important to emphasize that, since the MPD algorithm depends on evaluating a one dimensional but global optimization at each of its elementary steps, the algorithm may cycle and fail to reach the actual minimum of the LF. This translates into the expectation that the GD-based algorithms, originally designed for convex optimizations,  will outperform the MPD algorithm. Putting it differently, the MPD algorithm is designed to deal with the most difficult data fitting problems characterized by a very rugged landscape,  where more traditional GD-based algorithms fail to advance and are trapped in a local minimum. 

We plan to advance ideas put together in this manuscript in a number of directions, in particular we are working on (a) extending MPD algorithms to the case of Deep Learning by approximating the single-parameter message functions as PWL; (b) testing the MPD algorithm on other examples,  e.g. of Convolutional NN, Graphical NN, and other; (c) exploring other DF applications where locality of the training steps is expected to be a handicap, in particular of the type encountered in DL schemes  which are dominated by rare, and possibly adversarial, samples.

\section*{Acknowledgements}

This work was partially supported by M. Chertkov startup and LANL sub-contract to UArizona on "Machine Learning for Turbulence".   

\newpage

\appendix
\section{Details of the Message Passing Descent Algorithm}
\label{appendix}
 In the following, we present an explicit pseudo-code that implements the MPD Algorithm discussed in the main part (Section \ref{sec:piece-wise}) of the Manuscript. 

%At time $t=0$, we initialize parameters of the NN at random  in the Algorithm \ref{alg:initialization}.
\begin{algorithm}
\caption{Initialization}
\begin{algorithmic}[1] \label{alg:initialization}
\STATE  $\forall p\in \mathcal{V},\quad \theta_{p}\sim \mathcal{U}(-\sqrt{3\sigma},\sqrt{3\sigma})$.
\end{algorithmic}
\end{algorithm}
%\noindent The distribution $\mathcal{U}(-\sqrt{3\sigma},\sqrt{3\sigma})$ is the uniform distribution, with the zero mean and the variance, $\sigma$. If the MF $g[\bm{\theta}]$ in \eqref{eq:g-model} is a NN, a convenient choice for $\sigma$ is provided by the Kaiming initialization \cite{Kaiming}.
%After initialization, we run the MPD Algorithm for $T$ iterations.
%We brake the remainder of the MPD algorithm in three parts, constituting the main iterative step of MPD in Algorithm \ref{alg:MPD} and its two sub-routines presented in the {\bf Supplementary Materials}: Algorithm 3 -- subroutine MPD-SUM -- %\ref{alg:MPD-M}
%computing $\Theta_{p\to s}$ and $C_{p\to s}$, and Algorithm 4 -- subroutine GLOBAL-MIN -- %\ref{alg:GLOBAL-MIN} 
%executing Global one-dimensional optimization. 
\begin{algorithm}
\caption{MPD iteration:}
\textbf{Input:} Initial configuration $\bm{\theta}_p$, data-set $\mathbb{IO}^{(S)}$,\\
\textbf{Output:} $\bm{\theta}^*_p$.
\begin{algorithmic}[1] \label{alg:MPD}
\FOR{$t \in [T]$}
        \STATE Pick a vertex $p\in \mathcal{V}$: $p\sim U(\mathcal{V})$.
       \STATE 
       %$\forall s' \in \mathcal{F}/\{s\}$, compute $\Theta_{s'\to p}$ and $C_{s'\to p}$
       $\forall s \in \mathcal{F}$, compute $\Theta_{ s\to p}$ and $C_{s\to p}$ from $\bm{\theta}^{(t-1)}_{\partial {p_t}}$ and the IO sample $(\bm{x}^{(s)},\bm{y}^{(s)})$.
        \STATE Compute $\Theta_{p}$ and $C_{p}$ from $\{(\Theta_{s\to p},C_{s\to p})|\,s\in\mathcal{F}\}$. \\ \rightline{\COMMENT{Use subroutine MPD-SUM (Algorithm 3).}}
        \STATE  Update $\theta_{p}$ from $\Theta_{p}$ and $C_{p}$: $\theta_{p}\leftarrow \text{argmin}_{\widehat{\theta}_p}(h_{p}(\, \widehat{\theta}_p)\,).$\\ \rightline{\COMMENT{Use subroutine GLOBAL-MIN (Algorithm 4). Do not update other parameters.}}
%       \STATE  $\forall (p'\to s') \neq (p\to s),\quad  \theta_{p'\to s',t} =\theta_{p'\to s',t-1},$  \\ \rightline{\COMMENT{Do not update other parameters.}}
\ENDFOR
        \RETURN Result:  $\theta^*_p =\theta_p,\quad \forall p\in \mathcal{V}$
\end{algorithmic}
\end{algorithm}

In the following, e.g. in Algorithm \ref{alg:MPD-M}, we use the shortcut notation, $\bm{c}^{(r-1)}_{p}$, for the list of $Q+1$ coefficients of the function $h_{p}$ corresponding to the sub-domain $r$.
\begin{algorithm}
\caption{Subroutine MPD-SUM}
\textbf{Input:} $\Theta_{\partial s\to p},\quad C_{\partial s\to p}$\\
\textbf{Output:} $\Theta_{p},\quad C_{p}$ 
\begin{algorithmic}[1] \label{alg:MPD-M}
\STATE \textbf{Define}\  $\delta \bm{c}^{(r)}_{s\to p}=\bm{c}^{(r)}_{s\to p}- \bm{c}^{(r-1)}_{s\to p},\quad\forall s\in \mathcal{F},\, r \in [R]$.
\STATE \textbf{Define}\ $\mathcal{Z}=\quad\underset{\text{$1$st el.}}{\text{sort}}\bigcup_{s\in \mathcal{F}}\left(\left(\vartheta_{s\to p}^{(r)},\delta \bm{c}^{(r)}_{s\to p}\right)\Big|\, ,\, r\in [R]\right)$\footnotemark\\
\STATE \textbf{Define}\  $\mathcal{Z}_0=\left(\mathcal{Z}[1,1],\sum\limits_{s\in \mathcal{F}}\delta \bm{c}^{(1)}_{s\to p}\right)$\\
\rightline{\COMMENT{$\mathcal{Z}[1,1]$ is the first element of the first sub-list of $\mathcal{Z}$}}
\FOR {$r\in [S(R-1)]$}
\STATE \textbf{Define}\ $\vartheta^{(r)}_{p }= \mathcal{Z}[r,1]$\\
\STATE\textbf{Define}\ $\bm{c}^{(r)}_{p }=\bm{c}^{(r-1)}_{p}+\left(\mathcal{Z}[r,2],\cdots,\mathcal{Z}[r,Q+2]\right)$
\ENDFOR
\RETURN  $\Theta_{p}=\left\{\vartheta^{(r)}_{p }\big|r \in [SR]\right\}$ and $C_{p}=\left\{\bm{c}^{(r)}_{p }\big|r \in [SR]\right\}$
\end{algorithmic}
\end{algorithm}
%\footnotetext{
The function, $\underset{\text{$1$st el.}}{\text{sort}}(\,\cdot \,)$, in the line \#2 of the Algorithm \ref{alg:MPD-M}  sorts the elements of the list in the ascending order: 
\begin{gather*}
    \underset{\text{$1$st el.}}{\text{sort}}\bigcup_{s \in \mathcal{F}}\left(\left(\vartheta_{s\to p}^{(r)},\delta c^{(r,0)}_{s\to p},\cdots,\delta c^{(r,q)}_{s\to p}\right)\Big|r\in [R]\right)=\left(\left(\vartheta_{s_k\to p}^{(r_k)},\delta c^{(r_k,0)}_{s_k\to p},\cdots,\delta c^{(r_k,q)}_{s_k\to p}\right)\Big|k\in [SR]\right)%\\\qquad \text{with}\quad \forall k\leq k' \quad \vartheta_{s_{k}\to p}^{(r_k)}\leq \vartheta_{s_{k'}\to p}^{(r_{k'})}.
\end{gather*}%}
where, $\forall k\leq k':\ \vartheta_{s_{k}\to p}^{(r_k)}\leq \vartheta_{s_{k'}\to p}^{(r_{k'})}$.

The ordering operation in the line 2 of the Algorithm \ref{alg:MPD-M} is the most time-consuming operation, with the cost $O(R\times S \log(R \times S))$. The cost of evaluating the sum in the line 3 and of running line 4 is $O(Q\times R\times S)$. 

The global minimization step, $\theta^{(t)}_{ p_t}  =\underset{\theta_{p_t}'\in \mathbb{R}}{ \text{argmin}}\, h_{p_t}(\theta'_{p_t})$, see also Eq.~(23) or Eq.~(26) of the main manuscript, %\eqref{eq:global_min_t}
is obtained by, first, computing {\bf  analytically} the minimum in each of the sub-domains and then selecting the global minimum from the list of the sub-interval minima.
\begin{algorithm}
\caption{Subroutine GLOBAL-MIN}
\textbf{Input:}  $\Theta_{p},\quad C_{p}$\\
\leftline{\textbf{Output:} $\underset{\theta_p \in\mathbb{R}}{\text{argmin}}\,h_{p}(\theta_p),\quad \underset{\theta_p \in\mathbb{R}}{\text{min}}h_{p}(\theta_p)$ }
\begin{algorithmic}[1]\label{alg:GLOBAL-MIN}
\STATE $\forall r \in [R],\quad \epsilon^{(r)}=\underset{\theta_p \in[\vartheta^{(r)}_{p},\vartheta^{(r+1)}_{p}]}{\text{argmin}}\,h_{p}(\theta_p)$  
\STATE \textbf{Define}\  $\mathcal{E}=\left(\,\left(\epsilon^{(r)},\sum^Q_{q=0} c^{(r,q)}_{p}(\epsilon^{(r)})^q\right)\,\Big|\,r\in[R] \right)$
\RETURN  $\underset{\text{$2$nd el.}}{\text{min}}\,\mathcal{E}$
\end{algorithmic}
\end{algorithm}
\noindent Computational cost of the Alg. \ref{alg:GLOBAL-MIN} is proportional to the number of sub-domains in the $h$-functions %, e.g. $h_{p\to s}$, 
times the degree of the polynomial in the sub-domains (assumed the same in all the sub-domain), therefore resulting in the overall estimate, $O(Q\times S\times R)$.

According to the definition, $u_{s\to p}(\theta_p) =l_{s}( \theta_p,\theta^*_{\partial p})$, see also Eq.~(21) or Eq.~(24) of the main manuscript, % \eqref{eq:u_bp}, 
the (vectors of) coefficients,  $\Theta_{s\to p}$, $C_{s\to p}$, as well as  the number $R$, depend implicitly on the parameters $\theta^*_{\partial p}$,  therefore requiring re-computation at any step/update of the training process.

%\bibliography{example_paper}

\begin{thebibliography}{10}

\bibitem{2019Jose}
Casey Chu, Jose Blanchet, and Peter Glynn.
\newblock Probability functional descent: A unifying perspective on {GAN}s,
  variational inference, and reinforcement learning.
\newblock In Kamalika Chaudhuri and Ruslan Salakhutdinov, editors, {\em
  Proceedings of the 36th International Conference on Machine Learning},
  volume~97, pages 1213--1222, 2019.

\bibitem{UCI}
Dheeru Dua and Casey Graff.
\newblock {UCI} machine learning repository, 2017.

\bibitem{Kaiming}
K.~{He}, X.~{Zhang}, S.~{Ren}, and J.~{Sun}.
\newblock Delving deep into rectifiers: Surpassing human-level performance on
  imagenet classification.
\newblock In {\em 2015 IEEE International Conference on Computer Vision
  (ICCV)}, pages 1026--1034, 2015.

\bibitem{Javanmard-Montanari}
Adel Javanmard and Andrea Montanari.
\newblock State evolution for general approximate message passing algorithms,
  with applications to spatial coupling.
\newblock {\em Information and Inference}, 2, 11 2012.

\bibitem{data_set}
Manohar Kaul, Bin Yang, and Christian~S. Jensen.
\newblock Building accurate 3d spatial networks to enable next generation
  intelligent transportation systems.
\newblock In {\em Proceedings of the 2013 IEEE 14th International Conference on
  Mobile Data Management - Volume 01}, MDM '13, page 137–146, USA, 2013. IEEE
  Computer Society.

\bibitem{Adam}
Diederik Kingma and Jimmy Ba.
\newblock Adam: A method for stochastic optimization.
\newblock {\em International Conference on Learning Representations}, 12 2014.

\bibitem{VPM}
M~Mezard, G~Parisi, and M~Virasoro.
\newblock {\em Spin Glass Theory and Beyond}.
\newblock WORLD SCIENTIFIC, 1986.

\bibitem{pyThorch1}
Adam Paszke, Sam Gross, Soumith Chintala, Gregory Chanan, Edward Yang, Zachary
  DeVito, Zeming Lin, Alban Desmaison, Luca Antiga, and Adam Lerer.
\newblock Automatic differentiation in {PyTorch}.
\newblock In {\em NeurIPS Autodiff Workshop}, 2017.

\bibitem{pyThorch2}
Adam Paszke, Sam Gross, Francisco Massa, Adam Lerer, James Bradbury, Gregory
  Chanan, Trevor Killeen, Zeming Lin, Natalia Gimelshein, Luca Antiga, Alban
  Desmaison, Andreas Kopf, Edward Yang, Zachary DeVito, Martin Raison, Alykhan
  Tejani, Sasank Chilamkurthy, Benoit Steiner, Lu~Fang, Junjie Bai, and Soumith
  Chintala.
\newblock Pytorch: An imperative style, high-performance deep learning library.
\newblock In H.~Wallach, H.~Larochelle, A.~Beygelzimer, F.~d'Alch\'{e}Buc,
  E.~Fox, and R.~Garnett, editors, {\em Advances in Neural Information
  Processing Systems 32}, pages 8024--8035. Curran Associates, Inc., 2019.

\bibitem{Rangar}
S.~{Rangan}.
\newblock Generalized approximate message passing for estimation with random
  linear mixing.
\newblock In {\em 2011 IEEE International Symposium on Information Theory
  Proceedings}, pages 2168--2172, 2011.

\bibitem{NAG}
I.~Sutskever, J.~Martens, G.~Dahl, and G.~Hinton.
\newblock On the importance of initialization and momentum in deep learning.
\newblock {\em 30th International Conference on Machine Learning, ICML 2013},
  pages 1139--1147, 01 2013.

\bibitem{2005Yedidia}
J.~S. {Yedidia}, W.~T. {Freeman}, and Y.~{Weiss}.
\newblock Constructing free-energy approximations and generalized belief
  propagation algorithms.
\newblock {\em IEEE Transactions on Information Theory}, 51(7):2282--2312,
  2005.

\end{thebibliography}
%\bibliographystyle{icml2021}

%\bibliographystyle{plain}

\end{document}